\documentclass[10pt,twocolumn,letterpaper]{article}

\usepackage[pagenumbers]{iccv} 

\definecolor{iccvblue}{rgb}{0.21,0.49,0.74}
\usepackage[pagebackref,breaklinks,colorlinks,allcolors=iccvblue]{hyperref}
\usepackage{mathtools}

\def\paperID{2041} 
\def\confName{ICCV}
\def\confYear{2025}

\title{Global and Local Entailment Learning for Natural World Imagery}

\author{Srikumar Sastry, Aayush Dhakal, Eric Xing, Subash Khanal, Nathan Jacobs\\
Washington University in St.\ Louis\\
\{{\tt\small s.sastry, a.dhakal, e.xing, k.subash, jacobsn}\}{\tt\small @wustl.edu}
}

\begin{document}
\maketitle
\begin{abstract}
Learning the hierarchical structure of data in vision-language models is a significant challenge. Previous works have attempted to address this challenge by employing entailment learning. However, these approaches fail to model the transitive nature of entailment explicitly, which establishes the relationship between order and semantics within a representation space. In this work, we introduce Radial Cross-Modal Embeddings (RCME), a framework that enables the explicit modeling of transitivity-enforced entailment. Our proposed framework optimizes for the partial order of concepts within vision-language models. By leveraging our framework, we develop a hierarchical vision-language foundation model capable of representing the hierarchy in the Tree of Life. Our experiments on hierarchical species classification and hierarchical retrieval tasks demonstrate the enhanced performance of our models compared to the existing state-of-the-art models. Our code and models are open-sourced at \url{https://vishu26.github.io/RCME/index.html}.
\end{abstract}    
\section{Introduction}
\label{sec:intro}

Computer Vision has become increasingly valuable in understanding the natural world, thanks to the rise of open citizen science platforms and the abundance of consumer data. These tools, complemented by domain experts, have been instrumental in addressing pressing challenges at scale such as automatic species identification~\cite{vendrow2025inquire,van2018inaturalist}, animal behavior understanding~\cite{chen2023mammalnet, ng2022animal} and visual geolocalization~\cite{haas2024pigeon, vivanco2024geoclip}. Nevertheless, the complex and ever-changing nature of our world poses a significant challenge in constructing models that can generalize and adapt to novel data.

BioCLIP~\cite{stevens2024bioclip} and BioTroveCLIP~\cite{yang2025biotrove} successfully attempted to build a vision-language foundation model for the Tree of Life. Recently, TaxaBind~\cite{sastry2025taxabind} extended BioCLIP’s capabilities to handle additional modalities such as audio and satellite imagery. However, these models fail to fully leverage the hierarchical nature of the label space. This limited capability of these models limits them to reason at the most granular level of the hierarchy (i.e. \textit{species}) using a fixed database of taxonomic labels. Consequently, this restriction prevents the models from accurately representing the actual taxonomic system and the evolution of species in the Tree of Life.

We argue that learning hierarchical representations for the Tree of Life is crucial. A significant portion of species on Earth remain undescribed~\cite{pollock2025harnessing}, and labeling specimens up to the species rank is expensive and requires adequate expertise for biologists~\cite{gong2025clibd, rogers2023accelerating, su2021semi}. Furthermore, the taxonomic classification system and labels are subject to change over time due to mislabeling or the discovery of new species~\cite{stevens2024bioclip}. Hierarchical representations can allow for reasoning about such species at any rank and can eventually be used for grouping and routing specimens to biologists with appropriate expertise~\cite{pollock2025harnessing}. It can also help understand the evolution of certain species in the Tree of Life. For end users with arbitrary expertise, hierarchical representations facilitate classification at any taxonomic rank. Finally, in the paper, we empirically demonstrate the benefits of such structured representations for classification and retrieval tasks.

\begin{figure}[!t]
\begin{center}
   \includegraphics[width=\linewidth]{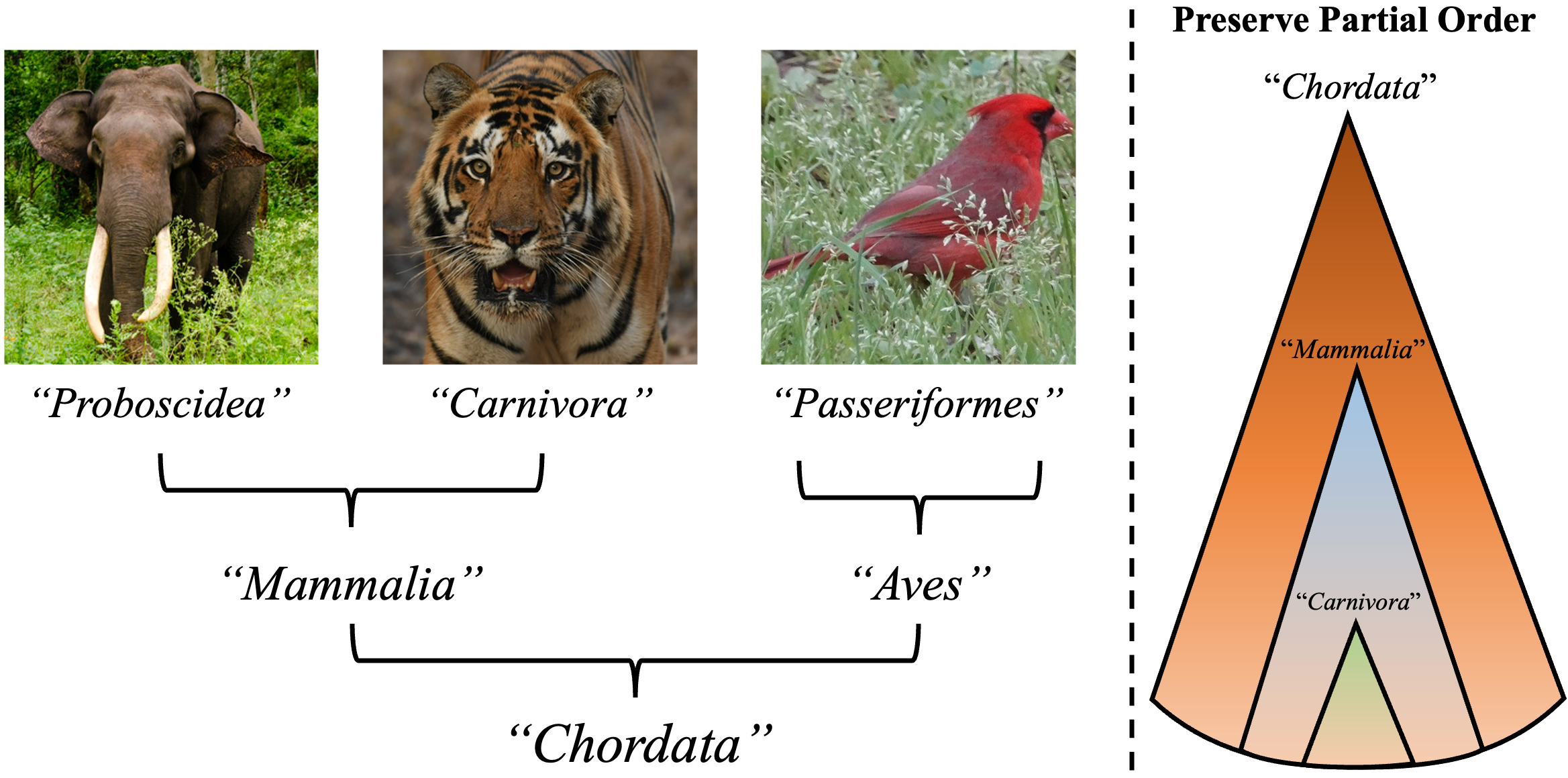}
\end{center}
   \caption{Conceptual overview of our method focusing on preserving the global order of concepts in vision-language models according to their distance from an entailment root. Our method aims to enforce transitivity in entailment.}
\label{fig:teaser}
\end{figure}

A popular technique for learning hierarchical representations for vision-language models is entailment learning, which aims to learn concentric cones of embedding subregions. In the past, studies on entailment learning relied on explicitly defining aperture angles that defined the structure of these cones~\cite{ganea2018hyperbolic,desai2023hyperbolic,pal2024compositional}. However, these approaches can be limiting since the optimality of the cone structures can vary from application to application. Recently, Alper~\etal~\cite{alper2024emergent} introduced radial embeddings, an approach to fine-tune existing vision-language models that eliminate the dependence of the objective on entailment cones. However, their method fails to enforce the partial order of concepts in their hierarchical embedding space.

In Figure~\ref{fig:teaser}, we illustrate transitivity in entailment, imposing a partial ordering of concepts in the embedding space~\cite{ganea2018hyperbolic}. For instance, if “\textit{Mammalia}” is entailed by “\textit{Chordata}” and “\textit{Carnivora}” is entailed by “\textit{Mammalia},” then “\textit{Chordata}” entails “\textit{Carnivora}”. Ideally, this phenomenon should hold for all possible sub-hierarchies in the data. Transitivity is an  important property for a representation space as it controls the distance between concepts based on their semantic granularity. For instance, fine-grained concepts are projected farther from coarse-grained concepts.

To this end, we propose a novel framework called Radial Cross-Modal Embeddings (RCME) which enables the learning of hierarchical representations by imposing partial order constraints while eliminating the need to define the structure of the cones. Using our framework, we propose a hierarchical vision-language foundation model for the Tree of Life, outperforming existing state-of-the-art models in hierarchical classification tasks. Notably, our framework is general enough to be adapted for any other domain. Our contributions are as follows:
\begin{enumerate}
    \item We propose an objective function to optimize for transitivity in textual entailment within vision-language models. We address the issue of Alper~\etal\cite{alper2024emergent}, which overlooks partial order in textual entailment.
    \item We propose Radial Cross-Modal Embeddings (RCME), a framework that solves for transitivity-enforced entailment and cross-modal alignment in vision-language models.
    \item Experiments show our models outperform the state-of-the-art in hierarchical classification, hierarchical retrieval, and image-to-image retrieval tasks.
    
\end{enumerate}

\begin{figure*}[!t]
\begin{center}
   \includegraphics[width=\linewidth]{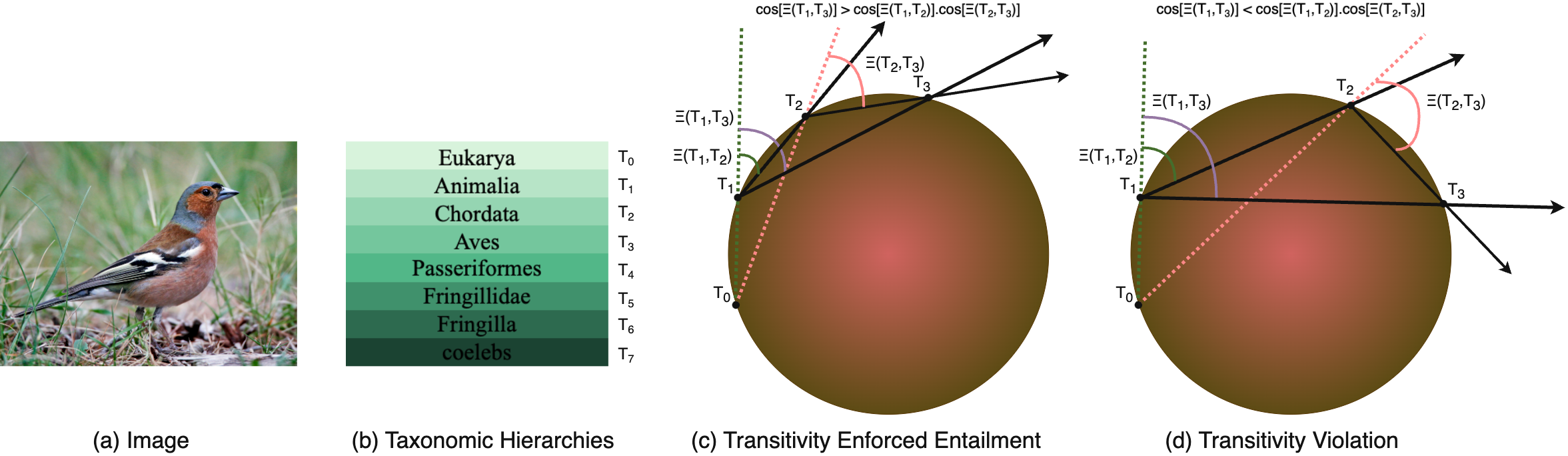}
\end{center}
   \caption{\textbf{Transitivity in Entailment}. In an ideal transitivity-imposed entailment, textual embeddings satisfy partial order conditions.}
\label{fig:entail}
\end{figure*}
\section{Related Works}
\subsection{Representation learning}
Contrastive learning has enabled training large-scale vision-language models \cite{radford2021learning, li2021align, jia2021scaling} which have generated significant advancement in diverse tasks like image classification and few-shot learning. Recent lines of work have focused on improving these generalist models by either achieving fine-grained alignment~\cite{yao2021filip, chen2023revisiting, li2022fine} or enhancing intra-modal representations~\cite{mistretta2025cross, chun2024probabilistic}. However, such methods are still not ideal for specific domains where there are structures of representations imposed by semantics. Hierarchical representation learning has gained traction, particularly in tasks requiring structured knowledge representation, such as natural language inference (NLI) ~\cite{le2019inferring, tifrea2018poincar, dhingra-etal-2018-embedding} and knowledge graph embeddings \cite{chami2020low, balazevic2019multi}. One of the key challenges in hierarchical representation learning is preserving the partial ordering of concepts in the embedding space while maintaining generalizability across different domains.

\subsection{Computer vision for ecology}
The intersection of computer vision and ecology has led to significant advances in tasks like fine-grained species classification~\cite{gong2025clibd, stevens2024bioclip}, animal detection using camera traps~\cite{simoes2023deepwild, beery2018recognition}, and animal behavior recognition~\cite{chen2023mammalnet, ng2022animal}. Large-scale datasets~\cite{maruf2024vlm4biobenchmarkdatasetevaluate, vendrow2024inquirenaturalworldtexttoimage, stevens2024bioclip, yang2025biotrove} and citizen science platforms like iNaturalist \cite{van2018inaturalist} have enabled the training of deep learning models to solve these tasks. Multimodal representation learning frameworks~\cite{sastry2024birdsat, huynh2024contrastive, daroya2024wildsat, chen2024mitree} with wildlife observations and satellite images has shown benefits in solving ecological tasks like species distribution modeling. 
Recently, vision-language foundation models for the Tree of Life such as BioCLIP \cite{stevens2024bioclip} and BioTroveCLIP~\cite{yang2025biotrove} have shown excellent capabilities in zero-shot species identification. TaxaBind \cite{sastry2025taxabind} extended such vision-language models by incorporating additional modalities such as audio and satellite imagery. However, all such models are limited to fixed taxonomic labels and struggle with classification at arbitrary taxonomic ranks. They lack structured hierarchical representations implied by the hierarchical nature of the Tree of Life.

\subsection{Entailment learning}
Traditional hierarchical learning approaches often rely on hyperbolic embeddings~\cite{nickel2017poincare, nickel2018learning, chami2019hyperbolic} to enforce hierarchical relationships. Entailment learning is particularly useful for structuring embeddings in a semantic order~\cite{vendrov2015order}. Early works on entailment learning explicitly defined cone structures using aperture angles to capture hierarchical dependencies~\cite{ganea2018hyperbolic, pal2024compositional, desai2023hyperbolic, yu2023shadow, wang2024learning}. However, these approaches are often restrictive as the optimal structure of the cones can vary across datasets and applications. Recent works, such as Radial Embeddings~\cite{alper2024emergent} and ATMG~\cite{ramasinghe2024accept}, attempt to relax these constraints by learning hierarchical embeddings without predefined cone structures in the radial and hyperbolic geometry respectively. While these methods improves adaptability, it does not explicitly enforce partial ordering, leading to suboptimal performance in hierarchical retrieval tasks. 
\section{Preliminaries}
We begin with the arguments on the conditions for entailment as proposed in Esteva \etal~\cite{esteva2012logics} and Ganea \etal~\cite{ganea2018hyperbolic}. For our purposes, we consider the tree of life hierarchy and develop the logic for entailment with respect to it. Let $\mathcal{R}_{\{j=0,1\ldots N\}}$ represent the sets of the domain of textual embeddings at each hierarchical rank. As $j$ increases, the semantic granularity in textual embedding increases. 
We consider $\mathcal{R}_0$ as the set which contains the entailment root embedding, $T_0$. Let $T_j^{i} \in \mathcal{R}_j$, denote the textual embedding for the $i^{\text{th}}$ species belonging to some rank $j$ in the hierarchy. We use $T_{j-1}^{i} \in \mathcal{R}_{j-1}$ to denote the immediate ancestor of $T_{j}^{i}$. 
Let $T_{j+1}^{i} \in \mathcal{R}_{j+1}$ represent the child of $T_{j}^i$. To define optimal radial cones, we first show the following.

\textbf{Lemma 1.} \textit{In a transitivity-enforced entailment, fine-grained concepts are progressively projected away from the entailment root and into smaller subregion when moving down in the hierarchy.}

Let $\mathfrak S_{T_j^i}$ and $\psi({T_j^i})$ denote a cone and its half aperture angle respectively defined at $T_j^i$ with respect to $T_0$. The transitivity property states that if $T_{j+1}^{i} \in \mathfrak S_{T_j^i}$, then $\mathfrak S_{T_{j+1}^{i}} \subseteq \mathfrak S_{T_j^i}$ ~\cite{ganea2018hyperbolic}. The direct consequence of this result is on the aperture angles of nested cones: $\psi({T_j^i}) \ge \psi({T_{j+1}^{i}})$. In other words, if $T_{j+1}^i$ is entailed by $T_{j}^{i}$, then the cone at $T_{j+1}^{i}$ is completely enclosed by the cone at $T_j^i$. This means that the distance of the embeddings from the root increases when one goes down in the hierarchy (see Equation~\ref{angle}). Hence, combining the above-stated results, fine-grained concepts (textual embeddings lower in the hierarchy) are contained within smaller cones than coarse-grained concepts. \textbf{We provide a mathematical proof in the appendix}. Lemma 1 is a natural property to have in textual entailment because it establishes a direct relationship between the semantic granularity of textual embeddings and their distance from the entailment root.

Ganea \etal\cite{ganea2018hyperbolic} defined the distance between two embeddings $T_j^{i}$ and $T_l^k$ in the entailment configuration as the exterior angle ($\Xi$) between $(T_j^{i}-T_0)$ and ($T_l^k - T_j^{i}$) considering the cone at $T_j^{i}$. In the Radial/Euclidean geometry, $\Xi$ is defined as follows:
\begin{equation}
    \Xi(T_j^{i}, T_l^k) = \arccos\left(\frac{\langle(T_j^{i}-T_0), (T_l^k - T_j^{i})\rangle}{||T_j^{i}-T_0||.||T_l^k - T_j^{i}||}\right)
\end{equation}
where $\langle\cdot{,}\cdot\rangle$ is the inner product between the embeddings. Likewise, the similarity measure $\mathcal{S}$ can be defined as:
\begin{equation}
    \mathcal{S}(T_j^{i}, T_l^k) = \cos(\Xi(T_j^{i}, T_l^k))
\end{equation}
 Furthermore, they defined the half aperture angle of any cone as a monotonically decreasing function with respect to its distance from the entailment root:
\begin{equation}
\label{angle}
    \psi(T_j^{i}) \propto \arcsin(1 / r(T_j^{i}, T_0))
\end{equation}
where $r$ is a distance function with range $[\epsilon, 1]$.

Esteva~\etal~\cite{esteva2012logics} proposed entailment configurations to adhere to the transitivity property that defines the partial order of concepts. Inspired from their formulation we devise the transitivity property mathematically as follows:
\begin{equation}
    \label{order}
    \mathcal{S}(T_{j-1}^{i}, T_{j+1}^{i}) \ge \mathcal{S}(T_{j-1}^{i}, T_{j}^{i}).\mathcal{S}(T_{j}^{i}, T_{j+1}^{i}) \:\forall\:j,i
\end{equation}
where $\mathcal{S}\in [0,1]$. This constraint establishes the relationship between text embeddings and their higher-level ancestors. In Figure~\ref{fig:entail}, we show two different scenarios in the entailment configuration. Figure~\ref{fig:entail}c) shows a perfect entailment configuration that satisfies transitivity constraints. In Figure~\ref{fig:entail}d) we show a configuration where transitivity is violated. 
Additionally, for entailment configuration where transitivity holds, Ganea \etal~\cite{ganea2018hyperbolic} showed that $\Xi(T_j^{i}, T_{j+1}^i)\le\psi(T_j^{i})\le\pi/2$ is true for any given parent and its child. This means that the cosine similarity between the two embeddings under transitivity constraints is always non-negative with respect to an entailment root.

Alper \etal~\cite{alper2024emergent} proposed vision-text radial embeddings by minimizing $\Xi$ for positive pairs while maximizing it for negative pairs. They perform text-only fine-tuning while keeping the vision encoder frozen. The main contribution was eliminating the dependence of the objective on aperture angle (Equation~\ref{angle}) and formulating the entailment problem on normalized embeddings. However, this resulted in the objective ignoring the transitivity constraint (equation~\ref{order}) and providing no guarantee that Lemma 1 holds. Their objective optimizes for the local entailment but does not necessarily solve for the partial order of concepts. Their objective is defined as follows:
\begin{equation}
\label{le}
    \mathcal{L}_{LE}(i, j, k) = \Xi(T_j^i, T_{j+1}^i) - \Xi(T_j^i, T_{j+1}^k)
\end{equation}
where $T_{j+1}^k$ is a negative example for $T_{j}^i$.
We argue that Alper~\etal~\cite{alper2024emergent} method only optimizes for the \textit{local entailment} objective, i.e. entailment with respect to the immediate ancestor.
\section{Method}
In this section, we introduce our proposed objective function, which seeks to optimize for transitivity without the requirement of defining an expression for the aperture angles. In addition, we describe our hard negative mining technique for improved performance. 
\subsection{Global Entailment Learning}
We begin by coining the terms local and global entailment. We say local entailment is enforced when $T_{j+1}^i$ is completely entailed by $T_{j}^i$ up to a reasonable degree, for all possible values of $i$ and $j$. Global entailment is enforced when Equation~\ref{order} holds for all possible sub-hierarchies in addition to local entailment. Mathematically, if $\mathcal{S}(T_j^i, T_{j+1}^{i}) = \gamma \:(\ge 0)$ and $\mathcal{S}(T_{j-1}^i, T_{j}^{i}) = \delta\:(\ge0)$, then $\mathcal{S}(T_{j-1}^i, T_{j+1}^{i}) \ge \gamma .\delta$. This ensures that Lemma 1 is satisfied. We enforce this objective using a margin-based loss as follows:
\begin{align}
\begin{split}
\label{ge}
    \mathcal{L}_{GE}(i, j;\alpha) &= \text{max}(0, \Xi(T_{j-1}^i, T_{j+1}^i) - \\
    & \arccos(\mathcal{S}(T_j^i, T_{j+1}^{i}).\mathcal{S}(T_{j-1}^i, T_{j}^{i})) +\alpha)
\end{split}
\end{align}
where $\alpha$ is the expected margin by which the angles should differ. We set it to the maximum possible value of $\pi/2$. We clip the values of $\mathcal{S}$ between $[0,1]$ for practical implementation. This loss is only calculated for consecutive positive triplets in a given hierarchy.

To enable global and local entailment learning, we combine the global and local objective functions. The loss is iteratively computed for each rank given positive and negative examples. The final loss is a combination of Equations~\ref{le} and~\ref{ge}:
\begin{align}
\begin{split}
\label{gle}
    \mathcal{L}_{GLE}(i, k;\alpha) &= \frac{1}{N-1}\underbrace{\sum_{p=1}^{N-1}\mathcal{L}_{GE}(i,p;\alpha)}_{\text{Global Entailment}} \\
    &+ \frac{1}{N}\underbrace{\sum_{p=0}^{N-1}\mathcal{L}_{LE}(i, p, K[p])}_{\text{Local Entailment}}
\end{split}
\end{align}
where $K$ represents a set of negative examples for each rank of the hierarchy, indexed by $p$. 

\begin{figure}[!t]
\begin{center}
   \includegraphics[width=\linewidth]{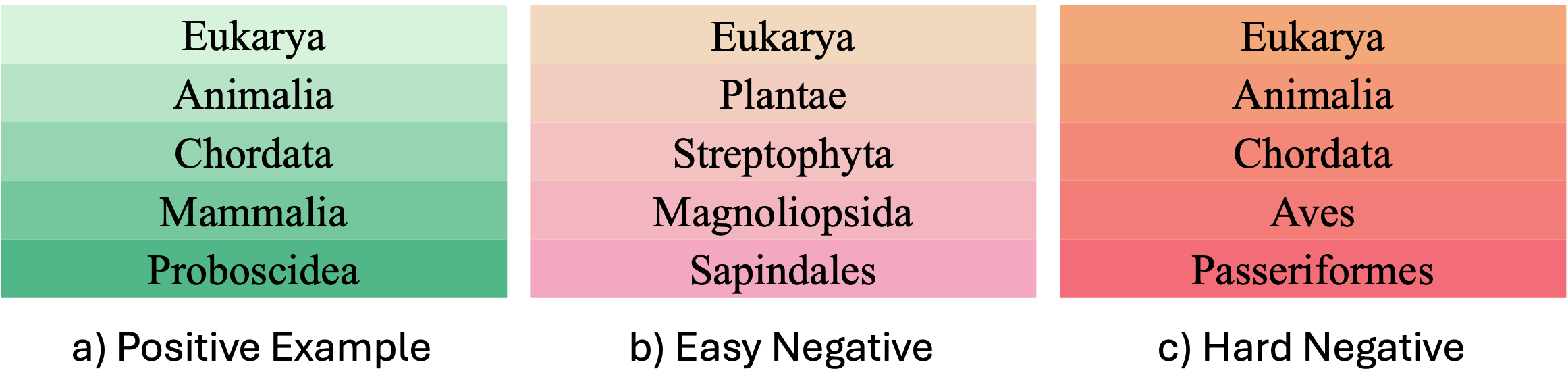}
\end{center}
   \caption{\textbf{Hard Negative Examples}. For the local entailment objective, we propose to sample negatives by matching all previous ranks of the positive examples. We recursively sample negatives for each rank separately.}
\label{fig:negative}
\end{figure}

\subsection{Radial Cross-Modal Embeddings}
In addition to our proposed global entailment objective, we also propose to add a cross-modal alignment loss term to fine-tune the vision encoder along with the text encoder. In \cite{alper2024emergent}, they added a prior preservation loss on the text encoder to preserve the original image-text embedding space. Instead, we propose to add a cross-modal alignment term to simultaneously fine-tune the vision and text encoders. This is given by:
\begin{equation}
\label{cma}
    \mathcal{L_\text{CMA}}(i) = - \log \frac{e^{\langle T_N^i, I^i\rangle}}{\sum_{m=1}^{B}e^{\langle T_N^m, I^m\rangle}+e^{\langle T_N^i,I^i\rangle}}
\end{equation}
where $I^i$ is an image embedding of the same species as represented by $T_N^i$. Note that the sum in the denominator is over a batch of $B$ negative samples. We only compute the objective for the most granular level of hierarchy which is the \textit{species} level. The final loss is now given by combining Equations~\ref{gle} and~\ref{cma}:
\begin{align}
\label{eq:rcme}
    \mathcal{L}_{\text{RCME}}(i, k;\alpha) = \mathcal{L}_{GLE}(i, k;\alpha) + \beta\mathcal{L_\text{CMA}}(i)
\end{align}
\subsection{Hard Negative Mining}
We propose a hard negative mining technique to sample negative examples required for the local entailment objective. Recall from equation~\ref{le} that the negative example must belong to the same rank as the positive example. To create a hard negative example for a given rank, we propose to sample labels that exactly match the taxonomic labels for all previous ranks of a given positive. In other words, we randomly sample a \texttt{sibling} of the parent for a given positive example. We then randomly sample a child of this sibling to create the final negative example. This is done recursively to create negative examples required at each rank. Figure~\ref{fig:negative} illustrates our hard negative sampling approach. This approach encourages the model to learn fine-grained differences between species of the same ancestry.
\section{Experiments}
In this section, we present the details of our implementation and the baselines used for comparison. We conduct four experiments to evaluate the effectiveness of the models on: 1) hierarchical retrieval and ordering of taxonomic labels; 2) zero-shot classification at each taxonomic rank; 3) intra-modal image-to-image retrieval at each taxonomic rank; 4) UMAP visualizations of textual embeddings. 

\subsection{Experimental Setup}

\textbf{Implementation Details}. We train two variants of our model. One model is trained using OpenCLIP's initialization and the other model is fine-tuned starting from the BioCLIP's checkpoint (denoted using FT). Both models are based on OpenCLIP's ViT-B/16 architecture. Both models are trained on the TreeofLife-10M dataset using 2 NVIDIA H100 GPUs. We use the word `\textit{Eukarya}' as the entailment root for the Tree of Life. Please refer to the appendix for additional details on the implementation.\\ 
\textbf{Baselines}. We compare our models against various vision-language baseline models, including CLIP~\cite{radford2021learning}, OpenCLIP~\cite{ilharco_gabriel_2021_5143773}, BioTroveCLIP~\cite{yang2025biotrove}, BioCLIP~\cite{stevens2024bioclip}, TaxaBind~\cite{sastry2025taxabind}, Radial Embeddings~\cite{alper2024emergent}, MERU~\cite{desai2023hyperbolic} and ATMG~\cite{ramasinghe2024accept}. Each of these models is based on the ViT-B/16 architecture. Since CLIP and OpenCLIP are not specifically trained for the task, we use the common names of species for image classification at the \textit{species} rank~\cite{stevens2024bioclip}. To ensure a fair comparison, we fine-tuned Radial Embeddings, MERU and ATMG on the TreeofLife-10M dataset, starting from BioCLIP’s checkpoint. Additionally, we do hard negative mining at each rank for fine-tuning.\\
\textbf{Evaluation Datasets}. We evaluate our models using the iNaturalist-2021~\cite{van2018inaturalist} and BioCLIP-Rare datasets~\cite{stevens2024bioclip}. The iNaturalist-2021 dataset comprises 10,000 unique species of animals, plants, and fungi. It includes a held-out test set with a total of 100,000 images. The BioCLIP-Rare dataset features 400 rare species of animals categorized under the IUCN Red List. Each species in the dataset is represented by 30 images for evaluation.

\subsection{Results}
\textbf{Ordering of taxonomic labels}. We evaluate the effectiveness of our learned vision-language representations on hierarchical retrieval tasks as defined in Alper~\etal~\cite{alper2024emergent} and Desai~\etal~\cite{desai2023hyperbolic}. Firstly, we check whether the taxonomic labels are correctly ordered according to their distance from the entailment root using Kendall's Tau ($\tau_d$). Secondly, we calculate image-to-text hierarchical retrieval metric relative to each taxonomic label. To ensure a fair evaluation, each unique species in the dataset is represented by a single image, which is selected at random from the test set. More details about the task setup present in the appendix.

Table~\ref{tab:order} presents the performance of the models on the iNaturalist-2021 dataset. Our learned representations exhibit excellent ordering of the taxonomic labels in the embedding space. This means embeddings corresponding to the \textit{species} rank are projected farthest from the root, while those corresponding to the \textit{kingdom} rank are projected closer to the root. Notably, radial embeddings perform worse than CLIP and OpenCLIP in the ordering task. Both our models show significant improvement, with the model trained using CLIP's checkpoint showing a more substantial performance improvement. We see a minimum absolute gain of +0.168 in correlation as compared to the baseline models. Furthermore, our method is able to outperform the rest of the methods in the hierarchical image-to-text retrieval task. We see a minimum absolute gain of +0.102 and +0.376 precision and recall respectively. Overall, these experiments demonstrate that our proposed objective function successfully imparts partial order to the embedding space.\\

\begin{table}[!t]
  \centering
  \begin{center}
  \resizebox{\columnwidth}{!}{%
  \begin{tabular}{lccccc}
    \toprule
    Model & Kendall's $\tau_d$ &Precision& Recall&F1\\
    \midrule
    CLIP~\cite{radford2021learning}&0.737&0.047&0.054&0.050\\
    OpenCLIP~\cite{ilharco_gabriel_2021_5143773}&\underline{0.825}&0.149&0.190&0.167\\
   \midrule
   BioTroveCLIP~\cite{yang2025biotrove}&0.566&0.122&0.173&0.143\\
   BioCLIP~\cite{stevens2024bioclip}&0.012&0.115&0.153&0.131\\
   TaxaBind~\cite{sastry2025taxabind}&0.012&0.116&0.155&0.133\\
   \midrule
   Radial Emb.~\cite{alper2024emergent}&0.521&0.147&\underline{0.196}&0.168\\
   MERU~\cite{desai2023hyperbolic}&0.403&\underline{0.356}&0.133&\underline{0.193}\\
   ATMG~\cite{ramasinghe2024accept}&0.571&0.343&0.130&0.189\\
   \midrule
   RCME$^{\text{FT}}$ (ours)&0.963&0.386&0.405&0.395\\
   RCME (ours)&\textbf{0.993}&\textbf{0.458}&\textbf{0.572}&\textbf{0.508}\\
    \bottomrule
  \end{tabular}
  }
  \caption{\textbf{Hierarchical Retrieval Metrics}. We evaluate the ability of different models to encode the partial order of taxonomies in the Tree of Life. Additionally, we evaluate the models on the standard task of hierarchical image-text retrieval.}
  \label{tab:order}
  \end{center}
\end{table}

\begin{table*}[!ht]
  \centering
  \begin{center}
  \begin{tabular}{lccccccc|c}
    \toprule
    Model & Kingdom &Phylum& Class&Family& Order & Genus&Species&Average\\
    \midrule
    CLIP~\cite{radford2021learning}&79.60&37.45&17.97&17.76&05.77&04.90&52.11&30.79\\
    OpenCLIP~\cite{ilharco_gabriel_2021_5143773}&66.72&18.42&15.45&07.80&02.60&04.42&58.55&24.99\\
   \midrule
   BioTroveCLIP~\cite{yang2025biotrove}&37.43&21.81&19.92&10.61&12.91&59.56&68.00&32.89\\
   BioCLIP~\cite{stevens2024bioclip}&36.96&32.02&19.97&24.31&31.43&61.04&68.24&39.13\\
   TaxaBind~\cite{sastry2025taxabind}&40.45&32.22&19.68&24.38&30.80&62.38&70.08&40.00\\
   \midrule
   Radial Emb.~\cite{alper2024emergent}&45.84&35.34&22.23&24.96&32.86&61.07&68.23&41.50\\
   MERU~\cite{desai2023hyperbolic}&\underline{95.82} &\underline{94.84} & \underline{63.12} & 27.27& 3.12& 1.30& 0.73&40.89\\
   ATMG~\cite{ramasinghe2024accept}&\textbf{99.12}&\textbf{86.79} &\textbf{73.03} &\textbf{51.83} &33.89& 49.59 &39.52&\underline{61.89}\\
   \midrule
   RCME$^{\text{FT}}$ (ours)&86.18&68.01&38.27&38.16&38.27&64.31&70.81&57.71\\
   RCME (ours)&88.18&84.81&55.22&\underline{46.74}&\textbf{41.82}&\textbf{67.41}&\textbf{73.52}&\textbf{65.09}\\
    \bottomrule
  \end{tabular}
  \caption{Zero-shot classification performance on iNaturalist-2021 dataset at various levels of the taxonomy.}
  \label{tab:zero-shot-inat}
  \end{center}
\end{table*}

\begin{table*}[!ht]
  \centering
  \begin{center}
  \begin{tabular}{lcccccc|c}
    \toprule
    Model &Phylum& Class&Family& Order & Genus&Species&Average\\
    \midrule
    CLIP~\cite{radford2021learning}&77.97&42.54&24.35&11.75&14.18&30.41&33.53\\
    OpenCLIP~\cite{ilharco_gabriel_2021_5143773}&20.35&33.56&19.14&04.42&10.54&30.22&19.71\\
   \midrule
   BioTroveCLIP~\cite{yang2025biotrove}&41.69&37.86&22.85&17.76&31.16&27.82&29.84\\
   BioCLIP~\cite{stevens2024bioclip}&43.55&61.08&53.25&45.32&53.38&34.52&48.51\\
   TaxaBind~\cite{sastry2025taxabind}&46.18&60.59&53.95&46.63&\underline{55.09}&\underline{35.84}&\underline{49.71}\\
   \midrule
   Radial Emb.~\cite{alper2024emergent}&43.77&63.03&53.96&45.75&53.43&34.85&49.13\\
   MERU~\cite{desai2023hyperbolic}&\underline{80.66}&58.99&25.92&08.12&05.13&04.53&30.56\\
   ATMG~\cite{ramasinghe2024accept}&\textbf{82.20}&\textbf{80.31}&\textbf{72.48}&\textbf{53.03}&45.02&35.32&61.39\\
   \midrule
   RCME$^{\text{FT}}$ (ours)&62.07&62.25&63.64&47.64&55.33&36.79&54.62\\
   RCME (ours)&79.60&\underline{77.34}&\underline{68.41}&\underline{50.10}&\textbf{56.66}&\textbf{41.62}&\textbf{62.64}\\
    \bottomrule
  \end{tabular}
  \caption{Zero-shot classification performance on BioCLIP-Rare dataset at various levels of the taxonomy. Note that this dataset primarily contains Animals.}
  \label{tab:zero-shot-rare}
  \end{center}
\end{table*}
\noindent
\textbf{Zero-shot classification}. We perform image classification by using taxonomic labels at each rank of the Tree of Life. The classification of each rank is performed independently to assess the ability of the models in the zero-shot setting. Table~\ref{tab:zero-shot-inat} and~\ref{tab:zero-shot-rare} show the performance of different models in this task on iNaturalist-2021 and BioCLIP-Rare datasets respectively. In both datasets, our model is able to exhibit excellent classification performance at each taxonomic rank. When averaging performance over each taxonomic rank, our model exhibits gains of +5.17\% and +2.03\% on iNaturalist-2021 and BioCLIP-Rare respectively. For radial emb., MERU and ATMG, we see a gain in performance at higher ranks of the taxonomy, while a dip in performance at fine ranks such as \textit{genus} and \textit{species}. This empirically demonstrates the usefulness of global entailment learning in preserving performance at fine ranks of the hierarchy. 

Interestingly, each of the models show lower performance for ranks \textit{class}, \textit{family}, and \textit{order} than for ranks \textit{genus} and \textit{species} in the iNaturalist dataset. This is because of the lower performance of the models for plants as compared to animals in these classes. Notably, this behavior is not seen in Table~\ref{tab:zero-shot-rare} since the dataset only contains animals. We suspect this happens as plants usually exhibit convergent traits and are usually mislabeled A combination of factors including morphological variability, hybridization, genetic complexity, and evolving methodologies usually complicates plant taxonomy. We show the kingdom-wise performance of our model on this dataset in the appendix to further analyze this behavior. This demonstrates the importance of learning hierarchical representations to detect such behavior and improve the taxonomic classification system.
\begin{table*}[!ht]
  \centering
  \begin{center}
  \begin{tabular}{lccccccc|c}
    \toprule
    Model & Kingdom &Phylum& Class&Family& Order & Genus&Species&Average\\
    \midrule
    CLIP~\cite{radford2021learning}&95.09&91.38&80.04&49.29&29.50&14.64&10.60&52.94\\
    OpenCLIP~\cite{ilharco_gabriel_2021_5143773}&96.29&93.33&85.42&59.85&41.85&26.11&16.78&59.95\\
   \midrule
   BioTroveCLIP~\cite{yang2025biotrove}&98.60&98.07&95.31&85.06&78.89&68.97&55.27&82.88\\
   BioCLIP~\cite{stevens2024bioclip}&98.50&97.60&94.94&84.98&78.63&67.22&51.62&81.92\\
   TaxaBind~\cite{sastry2025taxabind}&98.66&98.07&95.71&85.64&78.90&68.84&54.47&82.90\\
   \midrule
   Radial Emb.$^*$~\cite{alper2024emergent}&98.50&97.60&94.94&84.98&78.63&67.22&51.62&81.92\\
   MERU~\cite{desai2023hyperbolic}&98.20&96.02&85.04&58.81&41.96&26.56&16.40&60.43\\
   ATMG~\cite{ramasinghe2024accept}&\underline{99.16}&\underline{98.45}&\underline{96.88}&\underline{89.02}&\underline{82.30}&70.09&52.69&\underline{84.08}\\
   \midrule
   RCME$^\text{FT}$ (ours)&98.72&98.08&95.76&86.95&80.43&\underline{70.27}&\underline{57.10}&83.91\\
   RCME (ours)&\textbf{99.65}&\textbf{99.04}&\textbf{97.11}&\textbf{90.61}&\textbf{85.38}&\textbf{75.09}&\textbf{61.27}&\textbf{86.88}\\
    \bottomrule
  \end{tabular}
  \caption{\textbf{Image-to-Image Retrieval}. We evaluate the effectiveness of the intra-modal image representations learned by the models on the task of image-to-image retrieval at each taxonomic rank. $^*$Radial Emb. performs identically to BioCLIP since the vision encoder is not fine-tuned during its training.}
  \label{tab:im2im}
  \end{center}
\end{table*}

\begin{figure*}[!t]
\begin{center}
   \includegraphics[width=\linewidth]{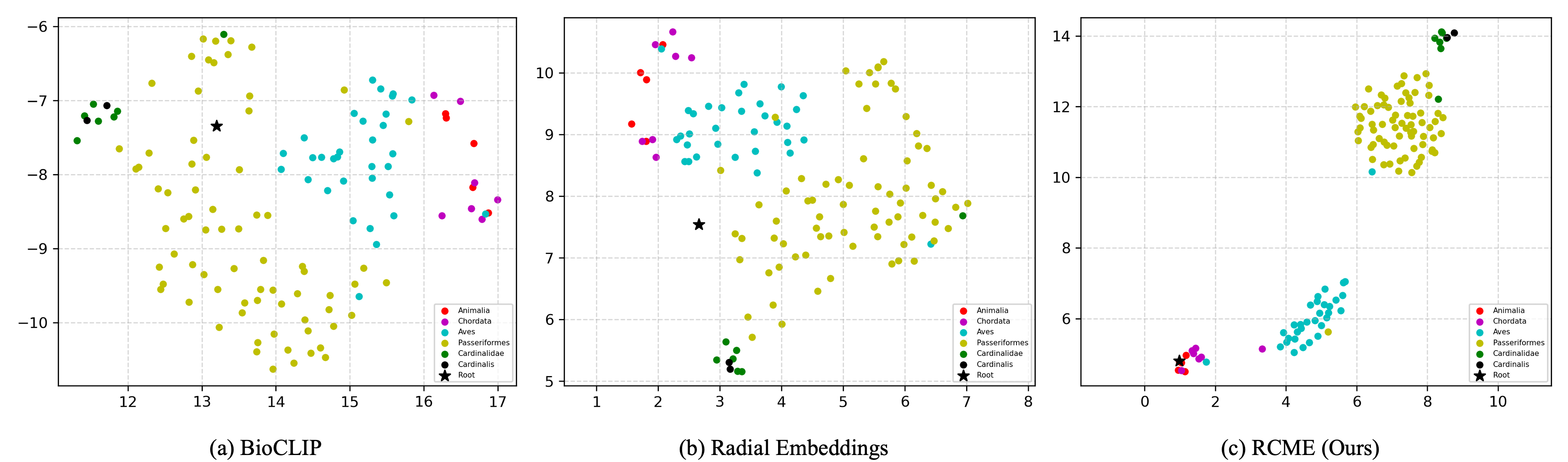}
\end{center}
   \caption{\textbf{UMAP Visualization of Textual Embeddings}. We visualize the textual embeddings using 2-D UMAP to show our model learns to preserve the partial order of taxonomic labels based on their distance from the entailment root.}
\label{fig:umap}
\end{figure*}
\noindent
\textbf{Image-to-image retrieval}. In this experiment, we explore whether our method enhances intra-modal representations. We conduct image-to-image retrieval at each taxonomic rank. Given an image of a particular species and its corresponding taxonomic label at a rank, our objective is to retrieve images of species with the same taxonomic label at the given rank. For example, given an image of an \textit{elephant} with the label \textit{Mammalia}, our goal is to retrieve images of \textit{Mammalia} using solely the image of the given \textit{elephant}.

Table~\ref{tab:im2im} presents the results of this experiment on the iNaturalist-2021 dataset. We compute the recall metric (R@1) for this task. Our method outperforms all other models under consideration. We see a gain of +3.33\% when performance is averaged over all taxonomic ranks. Notably, the radial embedding method’s performance is limited to the frozen image encoder model used, as it does not fine-tune the image encoder. There is a noticeable gap in the performance of the models at the \textit{species} rank as compared to our model showing that our model is effective at extracting fine-grained visual features. CLIP and OpenCLIP exhibit poor performance, particularly in the \textit{genus} and \textit{species} categories, indicating that these models struggle to extract fine-grained visual features. This suggests the potential of our model in future applications such as fine-grained retrieval augmented generation.

\begin{table*}[!ht]
  \centering
  \begin{center}
  \begin{tabular}{lcccccccccc|c}
    \toprule
    $\mathcal{L}_{LE}$ &$\mathcal{L}_{GE}$&$\mathcal{L}_{prior}$&$\mathcal{L}_{CMA}$ &Kingdom &Phylum& Class&Family& Order & Genus&Species&Average\\
    \midrule
    &&&\checkmark&36.96&32.02&19.97&24.31&31.43&61.04&68.24&39.13\\
    \checkmark&&\checkmark&&45.84&35.34&22.23&24.96&32.86&61.07&68.23&41.50\\
    
    \checkmark&&&\checkmark&47.34&38.14&25.78&26.88&35.34&62.67&69.43&43.65\\
    \checkmark&\checkmark&\checkmark&&85.13&81.11&53.21&44.33&39.82&65.90&71.28&62.97\\
    \checkmark&\checkmark&&\checkmark&\textbf{88.18}&\textbf{84.81}&\textbf{55.22}&\textbf{46.74}&\textbf{41.82}&\textbf{67.41}&\textbf{73.52}&\textbf{65.09}\\
    
    \bottomrule
  \end{tabular}
  \caption{\textbf{Loss Ablation}. We evaluate the performance of our model when trained with various combinations of the loss terms proposed in our objective function.}
  \label{tab:abla}
  \end{center}
\end{table*}

\begin{table}[!ht]
  \centering
  \begin{center}
  \resizebox{\columnwidth}{!}{%
  \begin{tabular}{lccc}
    \toprule
    RCME & iNaturalist-2021 &BioCLIP-Rare\\
    \midrule
   w/o negative mining&62.22&59.12\\
   with negative mining&\textbf{65.09}&\textbf{62.64}\\
    \bottomrule
  \end{tabular}
  }
  \caption{\textbf{Benefits of hard negative mining}. We evaluate the effectiveness of our hard negative mining approach for hierarchical representation learning of Tree of Life.}
  \label{tab:neg_sample}
  \end{center}
\end{table}

\noindent
\textbf{UMAP visualization of textual embeddings}. In Figure~\ref{fig:umap}, we present a 2-D Uniform Manifold Approximation and Projection (UMAP) visualization of textual embeddings obtained from BioCLIP, Radial Emb., and RCME. For a given \textit{species}, we plot the embedding for all the ancestors in the taxonomic hierarchy and their siblings. 
For instance, if we are given the specie \textit{Cardinalis cardinalis}, we begin with kingdom label \textit{Animalia} and plot all the \textit{Phyla} that belong to the kingdom \textit{Animalia}. This is repeated for all the subsequent ranks. From the plot, we anticipate two properties: 1) siblings share similar embeddings; 2) the embeddings are ordered in a coarse-to-fine manner (from kingdom to species), based on their distance from the entailment root. From the figure, it is evident that our model has successfully preserved the partial order of taxonomic labels based on their distance from the entailment root. For instance, the embeddings corresponding to the \textit{species} rank are projected farthest away from the entailment root. However, BioCLIP and Radial Emb. are unable to effectively enforce transitivity. We provide additional UMAPs in the appendix.

\subsection{Ablations}
We conduct an ablation study to analyze the effect of each loss component on the performance of our models. For the losses not using our cross-modal alignment term ($\mathcal{L}_{CMA}$), we include the $\mathcal{L}_{prior}$ from Alper~\etal~\cite{alper2024emergent} to preserve the original vision-language alignment. It is given as: $\mathcal{L}_{prior} = - \langle T_j^i, T_j^{i*} \rangle$, where $T_j^{i*}$ is an embedding from a frozen pre-trained text encoder. For losses using $\mathcal{L}_{prior}$, we perform fine-tuning starting from BioCLIP's checkpoint. Table~\ref{tab:abla} presents the performance of these losses for the image classification task on the iNaturalist-2021 dataset. Notably, incorporating our proposed global entailment objective function significantly enhances models’ performance compared to using only the local entailment objective. We notice a gain of +51.73\% in performance when adding our global entailment objective function to Alper~\etal's~\cite{alper2024emergent} objective function. Furthermore, our cross-modal alignment term outperforms the prior preservation loss. We notice a minimum gain of +3.36\% when replacing the prior preservation loss with our cross-modal alignment loss to train the vision and text encoder simultaneously.

We additionally investigate whether our proposed hard negative mining approach outperforms the random sampling approach. We evaluate the models trained using our negative mining and random sampling approaches on the iNaturalist-2021 and BioCLIP-Rare datasets. Table~\ref{tab:neg_sample} presents the comparison. We notice that we get a performance improvement of +4.61\% and +5.62\% on both datasets respectively. See appendix for additional ablations.

\begin{table}[!ht]
  \centering
  \begin{center}
  \resizebox{\columnwidth}{!}{%
  \begin{tabular}{lcccc}
    \toprule
    Model &Kendall's $\tau_d$&Precision&Recall&F1\\
    \midrule
    $\text{CLIP}^{\text{B}}$&0.883&\textbf{0.335}&0.142&0.199\\
    $\text{CLIP}^{\text{B}}$ (MERU)&0.855 &0.122& 0.401 &0.187\\
    $\text{CLIP}^{\text{B}}$ (ATMG)&0.981 &0.134 &0.422 &0.203\\
    $\text{CLIP}^{\text{B}}$ (HyCoCLIP)&0.892 &0.124 &\underline{0.451} &0.194\\
    $\text{CLIP}^{\text{B}}$ (Radial Emb.)&\underline{0.988}&0.155&0.441&\underline{0.229}\\
    $\text{CLIP}^{\text{B}}$ (RCME)&\textbf{0.991}&\underline{0.162}&\textbf{0.467}&\textbf{0.241}\\
    \midrule
    $\text{CLIP}^{\text{L}}$&0.881&\underline{0.151}&0.343&0.209\\
    $\text{CLIP}^{\text{L}}$ (Radial Emb.)&\underline{0.973}&0.145&\underline{0.415}&\underline{0.215}\\
    $\text{CLIP}^{\text{L}}$ (RCME)&\textbf{0.992}&\textbf{0.158}&\textbf{0.452}&\textbf{0.234}\\
    \bottomrule
  \end{tabular}
  }
  \caption{Hierarchical retrieval metrics on HierarCaps dataset. Our objective function results in improved ordering and image-to-text retrieval performance.}
  \label{tab:hierarcaps}
  \end{center}
\end{table}

\subsection{Generalization to HierarCaps}
To demonstrate the generalizability of our proposed objective function across other domains, we conducted experiments on the HierarCaps dataset~\cite{alper2024emergent}. This dataset comprises a subset of images from the Conceptual Captions (CC) dataset, each accompanied by captions at four different levels of granularity. We fine-tune the ViT-B/16 and ViT-L/14 variants of the CLIP model on this dataset using our proposed objective function in equation~\ref{eq:rcme} without hard negative mining. Once trained, we compute hierarchical retrieval metrics on the held-out test set of HierarCaps. As evident from Table~\ref{tab:hierarcaps}, our model outperforms radial embeddings in both ordering and hierarchical image-to-text retrieval tasks. These results demonstrate the successful application of our objective function in other application domains, enabling the imposition of a partial ordering along with entailment in an embedding space. 
\section{Conclusion}
In this work, we presented Radial Cross-Modal Embeddings (RCME), a framework for learning transitivity-enforced entailment in vision-language models. We proposed a novel objective function to enable global learning of entailment, which aids in preserving the partial order of concepts. Our framework not only improves cross-modal representations but also intra-modal representations. By leveraging our framework, we proposed a hierarchical foundation model for the Tree of Life, outperforming the state-of-the-art. We showed how hierarchical representations can improve the taxonomic classification of species and reveal unusual patterns in the taxonomic classification system, especially in plants. Our future works will focus on using the learned hierarchical representations to understand and comprehend species evolution, identify distinctive anomalies within the Tree of Life, and devise strategies to enhance the taxonomic classification system.
\section{Acknowledgements} 
This research used the TGI RAILs advanced compute and data resource which is supported by the National Science Foundation (award OAC-2232860) and the Taylor Geospatial Institute.
{\small
\bibliographystyle{ieeenat_fullname}
\bibliography{main}
}
\clearpage
\setcounter{page}{1}
\maketitlesupplementary

\appendix

\section{Proof for Lemma 1}
Lemma 1 states that finer-grained concepts are progressively projected: 1) away from the entailment root and 2) into smaller subregions in a transitivity-enforced entailment. We begin with the definition of distance in an entailment configuration:
\begin{equation}
    \Xi(T_j^{i}, T_l^k) = \text{arccos}\left(\frac{\langle(T_j^{i}-T_0), (T_l^k - T_j^{i})\rangle}{||T_j^{i}-T_0||.||T_l^k - T_j^{i}||}\right)
\end{equation}
where $\langle\cdot{,}\cdot\rangle$ is an inner product between the embeddings. The distance between two textual embeddings are computed with respect to the entailment root. In an entailment configuration with transitivity, the following property is satisfied~\cite{ganea2018hyperbolic} between a parent and its child:
\begin{equation}
    \Xi(T_j^{i}, T_{j+1}^i)\le\psi(T_j^{i})\le\pi/2
\end{equation}
This means that $\Xi(T_j^{i}, T_{j+1}^i) \in [0, \pi/2]$. It follows:
\begin{align}
\label{sims}
    0 \le \arccos\left(\frac{\langle(T_j^{i}-T_0),(T_{j+1}^i - T_j^{i})\rangle}{||T_j^{i}-T_0||.||T_{j+1}^i - T_j^{i}||}\right) \le \frac{\pi}{2}\\
    0 \le \left(\frac{\langle(T_j^{i}-T_0),(T_{j+1}^i - T_j^{i})\rangle}{||T_j^{i}-T_0||.||T_{j+1}^i - T_j^{i}||}\right) \le 1
\end{align}
Simplifying the above equation, we get the following expressions:
\begin{align}
    0 &\le \langle(T_j^{i}-T_0), (T_{j+1}^i - T_j^{i})\rangle\\
    0 &\le \langle T_j^{i}, T_{j+1}^i\rangle + \langle T_j^{i}, T_0\rangle - \langle T_{j+1}^i,T_0\rangle - \langle T_{j}^i,T_{j}^i\rangle
\end{align}

\paragraph{Case 1: \textit{Radial Geometry}}
In radial geometry, all textual embeddings lie on a unit hypersphere. As a result, the inner product between any two embeddings can never exceed the value of 1. As a result, we get the following expressions:
\begin{align}
    1 + \langle T_{j+1}^i, T_0 \rangle - \langle T_j^{i}, T_0 \rangle &\le \langle T_j^{i}, T_{j+1}^i \rangle\\
     1 + \langle T_{j+1}^i, T_0 \rangle - \langle T_j^{i}, T_0 \rangle &\le 1\\
      \langle T_{j+1}^i, T_0 \rangle - \langle T_j^{i}, T_0 \rangle &\le 0 \\
       \Aboxed {\langle T_{j+1}^i, T_0 \rangle &\le \langle T_j^{i}, T_0 \rangle}
\end{align}
As can be seen from equation 19 , the cosine similarity between $T_j^{i}$ and $T_0$ is always greater than that of $T_{j+1}^{i}$ and $T_0$. This means the distance of $T_{j+1}^{i}$ and $T_0$ is always greater than that of $T_{j}^{i}$ and $T_0$.

\paragraph{Case 2: \textit{Euclidean Geometry}}
In Euclidean geometry, the entailment root is considered to be the origin (a vector of zeros). This means $T_0=0$. Textual embeddings in this geometry are unnormalized and can have arbitrary norms. The distance of textual embeddings in this geometry is simply the L-2 norm. Using equation 6, we get the following expressions:
\begin{align}
    \langle T_{j}^i,T_{j}^i\rangle &\le \langle T_j^{i}, T_{j+1}^i\rangle\\
    ||T_{j}^i|| &\le ||T_{j+1}^i||.\cos\theta\\
    \Aboxed{||T_{j}^i|| & \le ||T_{j+1}^i||}
\end{align}
In Euclidean geometry, the norms of the embeddings increase with increasing ranks. 

In both geometries, we can conclude that \textit{the distance of textual embeddings monotonically increase with increasing ranks}. This leads to the following expression for the distance of an embedding from the root:
\begin{equation}
r(T_{j+1}^i, T_0) \ge r(T_j^i, T_0)
\end{equation}
\begin{equation}
\label{dist}
    r(T_j^i, T_0) = f(i, j; T_0)
\end{equation}
where $f$ is a monotonically increasing function with respect to the rank $j$ and $r$ is the distance function. Now let, the aperture angle of a cone defined at each textual embedding have the following expression (as done in~\cite{ganea2018hyperbolic}):
\begin{equation}
    \label{angle_supp}
    \psi(T_j^{i}) \propto \arcsin(1 / r(T_j^i, T_0))
\end{equation}
The above expression establishes the relation between the aperture angle of a cone defined at some textual embedding $T_j^i$ and its semantic granularity. From the expression, it is evident that the aperture angle monotonically decreases with increasing $j$ which defines its semantic granularity. Hence, we can conclude that fine-grained concepts are progressively projected into smaller subregions.\\\\
\textit{The proof is complete}.

\begin{table*}[!ht]
  \centering
  \begin{center}
  \begin{tabular}{lccccccc|c}
    \toprule
    Kingdom &\#~Samples &Phylum& Class&Family& Order & Genus&Species&Average\\
    \midrule
    Fungi&3410&68.09&38.24&31.40&23.78&63.84&73.05&49.73\\
    
    Plantae&42710&92.17&37.01&15.82&30.35&66.34&74.45&52.69\\
    Animalia&53880&84.73&72.86&73.40&55.68&68.25&70.41&70.89\\
    \bottomrule
  \end{tabular}
  \caption{Zero-shot classification performance for each distinct \textit{kingdom} class present in the iNaturalist-2021 dataset.}
  \label{tab:kingdom}
  \end{center}
\end{table*}

\section{Implementation Details}
All our models are based on the ViT-B/16 architecture and use the OpenCLIP implementation in PyTorch. For training, we use a learning rate of $1e^{-7}$ with OneCycleLR scheduler and the Adam optimizer. We use a batch size of 32 and accumulate gradient batches of 2. We use 2 NVIDIA H100 GPUs with the Distributed Data Parallel training strategy. We train for a single epoch. We found training for larger number of epochs hindered the performance of the model especially in the fine-grained taxonomic ranks like \textit{genus} and \textit{species}. We fixed the value of $\beta$ to $0.1$ and $1.0$ for the model trained from BioCLIP's~\cite{stevens2024bioclip} and OpenCLIP's~\cite{ilharco_gabriel_2021_5143773} checkpoints respectively. For our global entailment objective, we set the margin $\alpha$ to $\pi/2$.

\section{Experimental Setup}
Below we describe the details of the experiments done in the main paper.

\textbf{Ordering of taxonomic labels}. We use the same setup as Alper~\etal~\cite{alper2024emergent}. We sample 50 equally spaced points from the entailment root to the closest textual embedding to a given query image in the embedding space. At each point, we retrieve a textual embedding from a database which is closest to the given image embedding. We define a radius equivalent to the distance between the points for retrieving relevant embeddings at each level. We compute the Kendall's Correlation Coefficient ($\tau_d$) to evaluate the quality of ordinal association among the retrieved embeddings. Similarly, we compute precision and recall metric relative to the seven ranks of ground-truth taxonomic labels.

\textbf{Zero-shot image classification}. For each evaluation dataset, we first create a database of unique textual embeddings for each rank of the taxonomy. For a given rank, we compute the top-1 recall/accuracy metric on image to text retrieval task. Unlike the ordering task, we compute the accuracy metric for each taxonomic rank independently. From the experiments, we notice that the performance of the models does not decrease monotonically with increasing ranks of the taxonomy. In Table~\ref{tab:kingdom}, we present kingdom-wise performance of our model. We notice that classification performance of plants especially in the \textit{family} and \textit{order} ranks is abnormally low. We believe this is due to highly similar traits and mislabeling of plant species in these ranks. Note that in this experiment, we create independent database of textual embeddings for each kingdom.

\textbf{Image-to-image retrieval}. In this experiment, we retrieve images of species with a given taxonomic label at a given rank using a query image. For an evaluation dataset, we precompute the embeddings for each of the images. Subsequently, we retrieve images by calculating the cosine similarity between the query image embedding and the precomputed image embeddings. We compute the recall metric (R@1).

\begin{figure*}[!t]
\begin{center}
   \includegraphics[width=\linewidth]{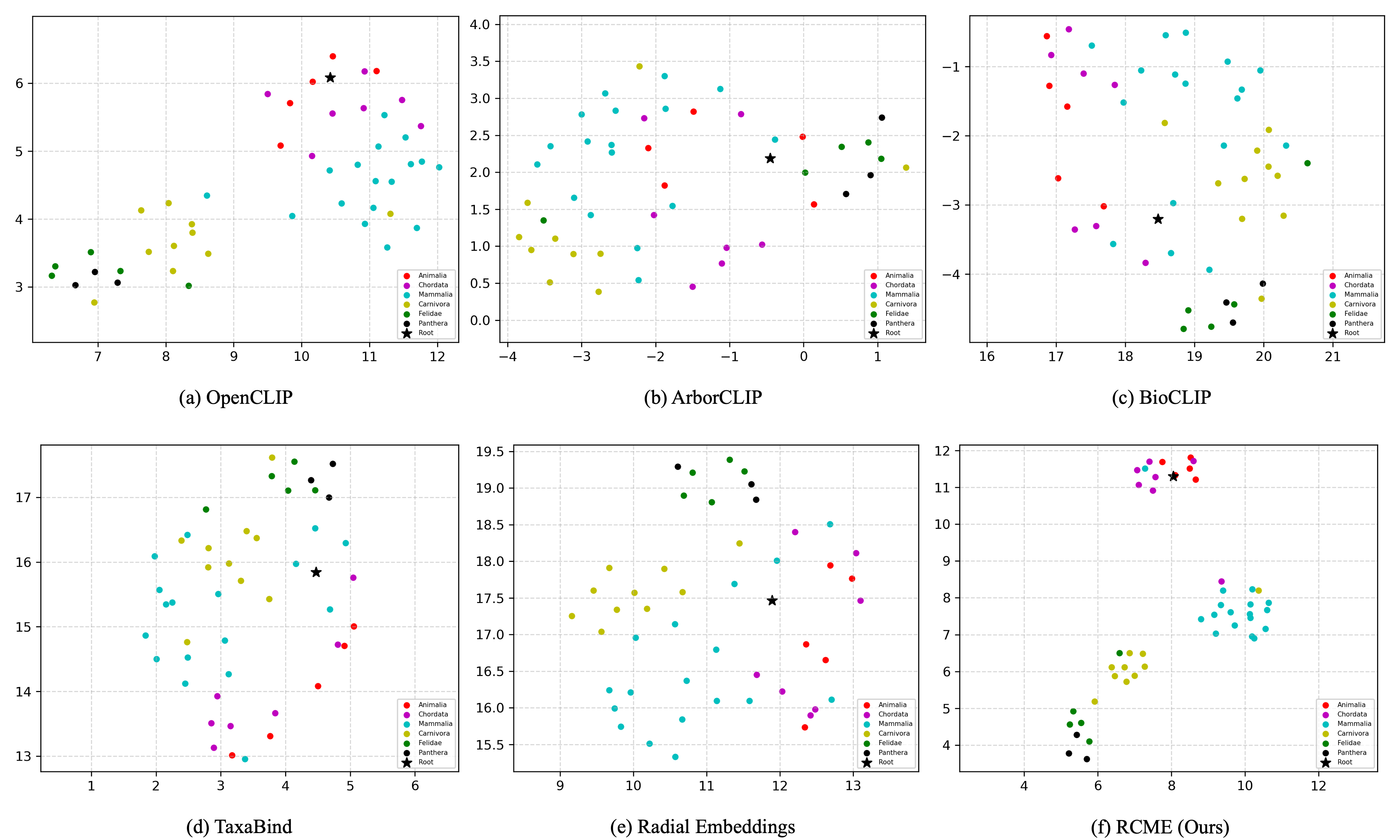}
\end{center}
   \caption{\textbf{UMAP Visualization of Textual Embeddings}. The visualizations show our model has successfully imparted partial order in the textual embeddings.}
\label{fig:umapsupp}
\end{figure*}

\textbf{UMAP visualization}. We show additional UMAP visualizations of textual embeddings from the models in Figure~\ref{fig:umapsupp}.

\section{Additional Ablations}
In Table~\ref{tab:alpha}, we show the performance of our global objective function with varying margins (see equation 6 in the main paper). We see that our objective function's performance improves with increasing margins.

\begin{table}[!ht]
  \centering
  \begin{center}
  \resizebox{\columnwidth}{!}{%
  \begin{tabular}{lcccc}
    \toprule
    $\alpha$ &Kendall's $\tau_d$&Precision&Recall&F1\\
    \midrule
    $\pi/2$&0.991&0.162&0.467&0.241\\
    $\pi/4$&0.990&0.152&0.467&0.229\\
    $\pi/8$&0.990&0.154&0.470&0.232\\

    $0$&0.990&0.151&0.454&0.226\\
    \bottomrule
  \end{tabular}
  }
  \caption{Hierarchical retrieval metrics on HierarCaps dataset with varying margins ($\alpha$) in our global entailment objective.}
  \label{tab:alpha}
  \end{center}
\end{table}
Additionally, we assess our model’s performance in the ordering task by varying the number of retrieval steps in the embedding space. Tables~\ref{tab:steps_inat} and~\ref{tab:steps} present the results. Reducing retrieval steps improves precision, but negatively affects recall. The ordering performance remains consistent, as expected.

\begin{table*}[!t]
  \centering
  \begin{center}
  \resizebox{\columnwidth}{!}{%
  \begin{tabular}{lcccc}
    \toprule
    Steps &Kendall's $\tau_d$&Precision&Recall&F1\\
    \midrule
    10&0.993&0.527&0.472&0.498\\
    20&0.993&0.491&0.552&0.520\\
    30&0.993&0.493&0.618&0.548\\
    40&0.993&0.455&0.568&0.505\\
    50&0.993&0.458&0.572&0.508\\
    \bottomrule
  \end{tabular}
  }
  \caption{Hierarchical retrieval metrics on iNaturalist-2021 dataset with varying number of retrieval steps.}
  \label{tab:steps_inat}
  \end{center}
\end{table*}

\begin{table*}[!t]
  \centering
  \begin{center}
  \resizebox{\columnwidth}{!}{%
  \begin{tabular}{lcccc}
    \toprule
    Steps &Kendall's $\tau_d$&Precision&Recall&F1\\
    \midrule
    10&0.991&0.224&0.344&0.271\\
    20&0.991&0.190&0.419&0.261\\
    30&0.991&0.174&0.450&0.251\\
    40&0.991&0.165&0.465&0.244\\
    50&0.991&0.162&0.467&0.241\\
    \bottomrule
  \end{tabular}
  }
  \caption{Hierarchical retrieval metrics on HierarCaps dataset with varying number of retrieval steps.}
  \label{tab:steps}
  \end{center}
\end{table*}

\end{document}